\documentclass[11pt,a4paper]{article}
\usepackage[hyperref]{acl2020}
\usepackage{times}
\usepackage{latexsym}
\usepackage{times}
\usepackage{latexsym}
\usepackage{graphicx}
\usepackage{url}
\usepackage{algorithm}
\usepackage{algpseudocode}
\usepackage{amsmath}
\usepackage{amssymb}
\usepackage{multirow}
\usepackage{verbatim}
\usepackage{times}
\usepackage{latexsym}
\usepackage{times}
\usepackage{latexsym}
\usepackage{graphicx}
\usepackage{url}
\usepackage{algorithm}
\usepackage{makecell}
\usepackage{algpseudocode}
\usepackage{amsmath}
\usepackage{amssymb}
\usepackage{multirow}
\usepackage{verbatim}

\usepackage{microtype}

\aclfinalcopy % Uncomment this line for the final submission
\def\aclpaperid{795} %  Enter the acl Paper ID here

\title{
Evidence-Aware Inferential Text Generation with \\Vector Quantised Variational AutoEncoder 
}
 
\author{Daya Guo$^1$\thanks{\ \ \ Work done while this author was an intern at Microsoft Research.} , Duyu Tang$^2$, Nan Duan$^2$, Jian Yin$^1$, Daxin Jiang$^3$ and Ming Zhou$^2$\\
	$^1$ The School of Data and Computer Science, Sun Yat-sen University.\\
	Guangdong Key Laboratory of Big Data Analysis and Processing, Guangzhou, P.R.China\\
	$^2$ Microsoft Research Asia, Beijing, China \\
	$^3$ Microsoft Search Technology Center Asia, Beijing, China \\
	{\tt \{guody5@mail2,issjyin@mail\}.sysu.edu.cn}\\
	{\tt \{dutang,nanduan,djiang,mingzhou\}@microsoft.com}\\
}

\date{}

\begin{document}
\maketitle
\begin{abstract}
%We study the problem of ...., which is to generate multiple ... for a given event. 
%Existing works typically only take the event as the input, while failing the background knowledge, which is crucial for providing useful clues for geenrationg outcomes.
%To addreass this problem, we propose a ... with ..

Generating inferential texts about an event in different perspectives requires reasoning over different contexts that the event occurs.
Existing works usually ignore the context that is not explicitly provided, resulting in a context-independent semantic representation that struggles to support the generation. 
To address this, we propose an approach that 
% , short for evidence-aware vector quantised variational autoencoder.
%, an evidence-aware approach for inferential text generation.
% It 
automatically \mbox{finds} evidence for an event from a large text corpus, and leverages the evidence to guide the generation of inferential texts. 
Our approach works in an encoder-decoder manner and is equipped with  a Vector Quantised-Variational Autoencoder, where the encoder outputs representations from a distribution over discrete variables. 
Such discrete representations 
%This 
enable automatically selecting relevant evidence, which not only facilitates evidence-aware generation, but also provides a natural way to uncover rationales behind the generation.
Our approach provides state-of-the-art performance on both Event2Mind and ATOMIC datasets.
More importantly, we find that with discrete representations, our model 
% can be used to 
\mbox{selectively} uses  evidence to generate different inferential texts.
\end{abstract}

\section{Introduction}
\begin{figure}[t]
	\centering
	\includegraphics[width=.48\textwidth]{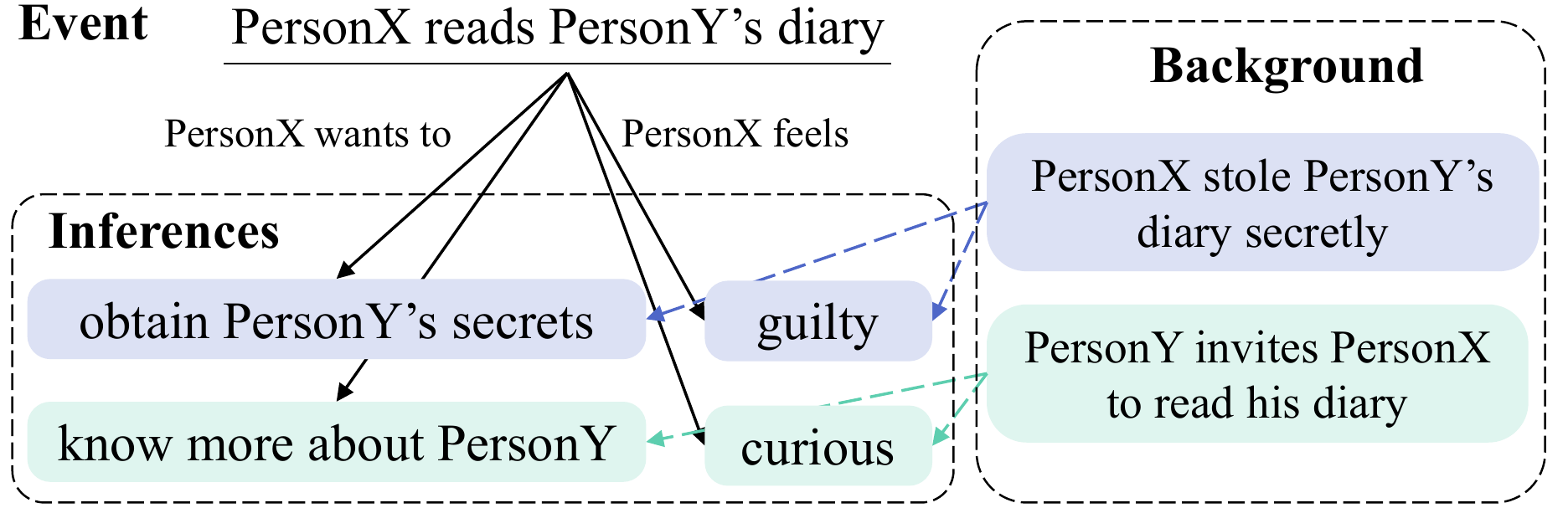}
	\caption{An examples of inferential text generation on mental states of event participants. We show two kinds of reasonable inferences for the event under different background knowledge that is absent in the dataset.}
	\label{fig:example}
\end{figure}

Inferential text generation aims to understand daily-life events and generate texts about their underlying causes, effects, and mental states of event participants, which is crucial for automated commonsense reasoning. 
% For example, in Figure \ref{fig:example}
Taking Figure \ref{fig:example} as an example, given an event \textit{``PersonX reads PersonY's diary''}, the cause of the participant ``\textit{PersonX}'' is to ``\textit{obtain Person Y's secrets}'' and the mental state of ``\textit{PersonX}'' is ``\textit{guilty}''. 
Standard approaches for inferential text generation \cite{rashkin2018event2mind,sap2019atomic,Bosselut2019COMETCT,du2019modeling} typically only take the event as the input, while ignoring the background knowledge that provides crucial evidence to generate reasonable inferences. 
% However, when a daily event occurs, human usually reason about its causes and effects under a particular background. 
% Taking Figure \ref{fig:example} as an example, given an event \textit{``PersonX reads PersonY's diary''}, 
For example, if the background knowledge of this example is ``\textit{PersonY invites PersonX to read his diary}'', the outputs should be different.
% outputs. 
% Taking Figure \ref{fig:example} as an example, given an event \textit{``PersonX reads PersonY's diary''}, the background \textit{``PersonX stoles PersonY's diary secretly''} can help the human reason about the motivation  of the event participant \textit{``PersonX wants to obtain PersonY's secrets''}.
%background knowledge can help the model reason whether the feeling of the event participant \textit{``PersonX''} is positive (e.g. curious) or negative (e.g. guilty).
%the plausible motivation and feeling of the event participant such as \textit{``PersonX''} may be various under different background. 
% Therefore, background knowledge is important for providing useful information to generate inferential texts of an event.

% There still remains a challenging problem. As illustrated in Figure \ref{fig:example}, the plausible motivation and feeling of the event participant may be various under different background, which require a way to select relevant evidence as supporting background knowledge for different inference.
%The task is challenging because the task calls for exploiting background knowledge to generate multiple reasonable inferences.

%task to test system’s ability on the reasoning of commonsense knowledge in NLP. 
%Given an event, the task of inferential text generation aims to reason about its causes, effects, and mental states of event participants. 

% In this work, we study how to automatically find relevant evidence as the supporting background knowledge and leverage the evidence to facilitate inferential text generation.

In this paper, we present an evidence-aware generative model, which 
% first finds relevant evidence as the supporting background knowledge and leverage the evidence to facilitate inferential text generation.
% To address the problem, we propose an evidence-aware vector quantised variational autoencoder (\modelname),
first retrieves relevant evidence from a large text corpus and then leverages retrieved evidence to guide the generation of inferential texts. 
Our model is built upon Transformer-based \cite{vaswani2017attention} encoder-decoder architecture, and is equipped with Vector Quantised-Variational Autoencoder to map an event to a discrete latent representation \cite{van2017neural}.
% fromdiscrete hidden representations.
%These discrete latent representations embody the semantics for generating from different perspectives, each of which stands for the generation with one kind of background knowledge, 
%thus supporting
% of making generation for a given event, so that they can support
%more controllable sampling and selective use of evidence for generating texts 
% with 
 %from different background knowledge.
%  (i.e. with
% different evidence.
% be used to selectively 
% In addition, 
%Furthermore, 
% our model not only supports 
% % and each event is mapped to a representations from a distribution over discrete variable.
% % This not only supports 
% efficient and more controllable sampling over different perspectives for generation, 
% , which benefits subsequent  evidence retrieval and text generation steps, 
% enables us to conveniently sample different semantics for finding evidence,
These discrete representations embody the latent semantic distribution of inferences given the event, thus supporting selection of relevant evidence as background knowledge to guide the generation in different perspectives.
Furthermore, 
% our model not only supports 
% % and each event is mapped to a representations from a distribution over discrete variable.
% % This not only supports 
% efficient and more controllable sampling over different perspectives for generation, 
% , which benefits subsequent  evidence retrieval and text generation steps, 
% enables us to conveniently sample different semantics for finding evidence,
% manipulate the s efficient samp, 
our model has two attractive properties: (1) it avoids the problem of posterior collapse, caused by latent variables being ignored, in traditional variational autoencoder with continuous latent variables
% with continuous hidden representations 
\cite{van2017neural}, and more importantly (2) it uncovers the rationale of a generation to some extent through tracing back the evidence that guides the generation and the selected discrete representation of the event.

We evaluate our approach on Event2Mind \cite{rashkin2018event2mind} and ATOMIC \cite{sap2019atomic} datasets, both of which focus on reasoning about causes and effects of events and mental states of event participants. 
Experimental results show that our approach achieves state-of-the-art performances on both datasets.
Further analysis shows that our approach can equip the generation with an explicit control over the semantics of latent variables and selected evidence to generate inferential texts in different perspective. The source codes are available at \url{https://github.com/microsoft/EA-VQ-VAE}.
%Further analysis show that both the context-aware retriever and the meta-learning strategy improve the performance.

%However, the task is still challenging for NLP systems, because it require background knowledge to reason. 

\section{Task Definition and Datasets}
Figure \ref{fig:example} shows an example of the task, which aims to generate inferential texts about causes and effects of daily-life events and mental states of the event’s participants. Formally, given an event $x=\{x_1,x_2,..,x_n\}$ and an inference dimension $r$ such as causes of the event, the goal is to generate multiple inferential texts $Y=\{y^{(1)},y^{(2)},...,y^{(m)}\}$\footnote{We use \textit{inference} and \textit{inferential text} interchangably}, where the background knowledge of the event is absent in the dataset.

We conduct experiments on Event2Mind\footnote{\url{https://uwnlp.github.io/Event2Mind/}} \cite{rashkin2018event2mind} and ATOMIC\footnote{\url{https://homes.cs.washington.edu/~msap/ATOMIC/}} \cite{sap2019atomic} datasets.
Both datasets contain about 25,000 unique events extracted from multiple data sources and provide multiple inferences under different inference dimensions by crowd-sourcing on Amazon Mechanical Turk. Event2Mind and ATOMIC contain 2.6 and 3.6 inferences on average per example, respectively. Event2Mind focuses on three inference dimensions related to mental states of participants (i.e. intents and reactions of the event’s participants), while ATOMIC has broader inference dimensions including mental states, probable pre- and post conditions of the event, and persona status.
More details about the two datasets are provided in the Appendix A. 
% (e.g., ROC Story~\cite{P18-1043}). 
%Event2Mind focuses on three relations related to mental states (i.e., intents and reactions of the actors), while ATOMIC has broader inferential dimensions includes mental states (the mental pre- and post- conditions of events), event (events about pre- and post- conditions of events) and persona (a stative relation about how the subject of an event is perceived).
%ATOMIC:3.6 Event2Mind:2.6

\section{Overview of the Approach}
We present our approach in this section, which first retrieves relevant evidence from a large text corpus, and then utilizes retrieved evidence as background knowledge to generate inferences.

\begin{figure}[t]
	\centering
	\includegraphics[width=0.48\textwidth]{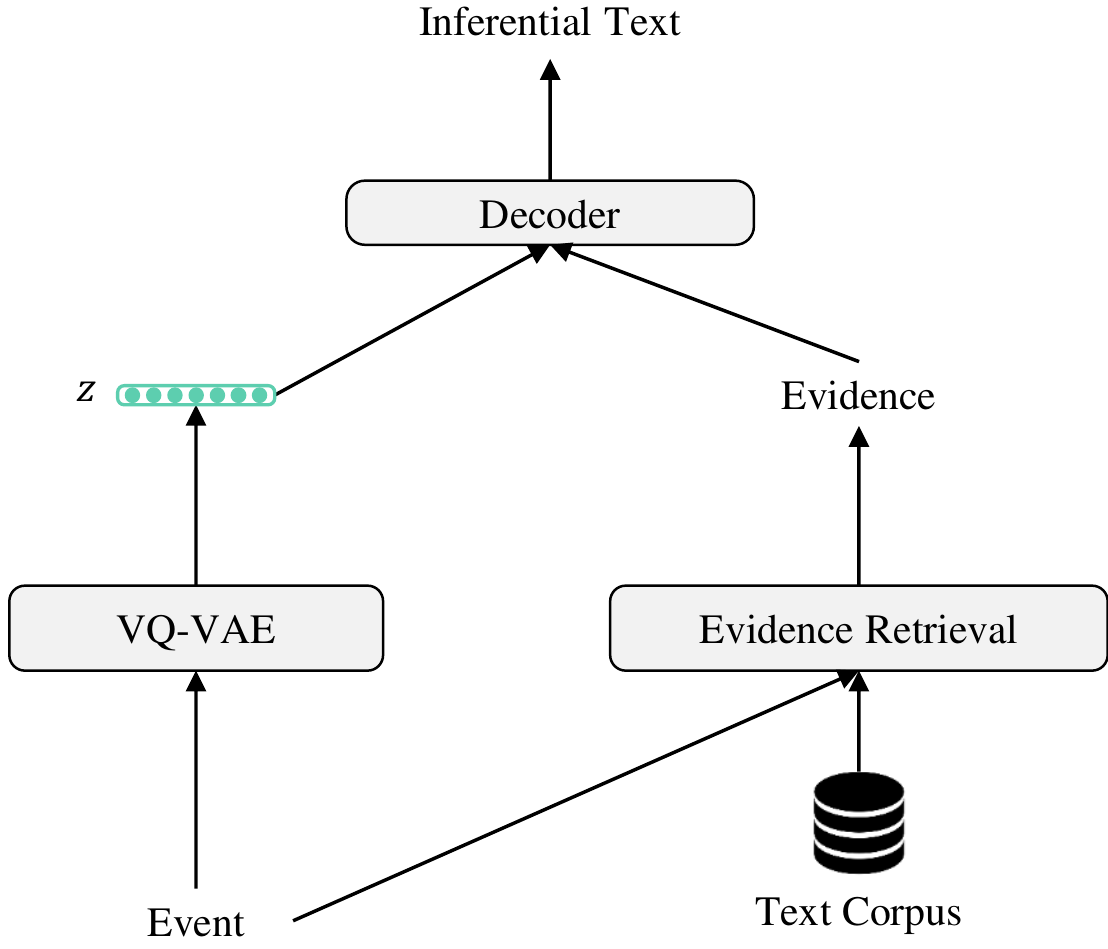}
	\caption{An overview of our approach.}
	\label{fig:model}
\end{figure}

Figure \ref{fig:model} gives an overview of our approach. 
First, our encoder takes an event as the input and outputs a semantic representation $z$ from a distribution over discrete latent variables, which is based on Vector Quantised-Variational Autoencoder (VQ-VAE) \cite{van2017neural}.
We then use the event as a query to retrieve top $K$ evidence from a large text corpus as background knowledge.
Lastly, the evidence-aware decoder takes the semantic representation and evidence as the input and generates the inference $y$, where the semantic representation selectively uses relevant evidence as background knowledge to guide the generation of inferences. 

%In order to enable latent variables to capture semantic content, we first train VQ-VAE by reconstructing the target. Then, the semantic latent variable $z$ is used to extract relevant evidence $c$ from a large-scale text corpus guide the generation. In the generation, we use the transformer language model $p_m(y|x,r,c)$ as our decoder, which takes the extracted evidence $c$ and the event $x$ with a relation $r$ as the input and generate the inference $y$.

%which equips the model with an explicit control over latent variables

%In the next, we will describe VQ-VAE, the way to retrieve evidence, evidence-aware decoder, and the method to effectively train our model one by one. 

%The latent variable is used to extract a related evidence $c$ from a large-scale text corpus to guide the generation. 
%Specially, we take the events as a query to search candidate evidence by the Elastic Search engine\footnote{\url{https://www.elastic.co/}}. 

%we first pre-train a VQ-VAE that can encode an example into a latent variable $z$, and then use the latent variable to select a related evidence from candidate evidence as background knowledge. Lastly, the transformer language model $p_m(y|x,r,c)$ takes the extracted evidence $c$ and the event $x$ with a relation $r$ as the input, and generate the inference $y$. 

%The details about the \modelname and the transformer language model will be introduced in Sections \ref{sec:selector} and Section \ref{sec:reasoner}, respectively.

\subsection{Vector Quantised-Variational Autoencoder}
\begin{figure*}[t]
	\centering
	\includegraphics[width=0.90\textwidth]{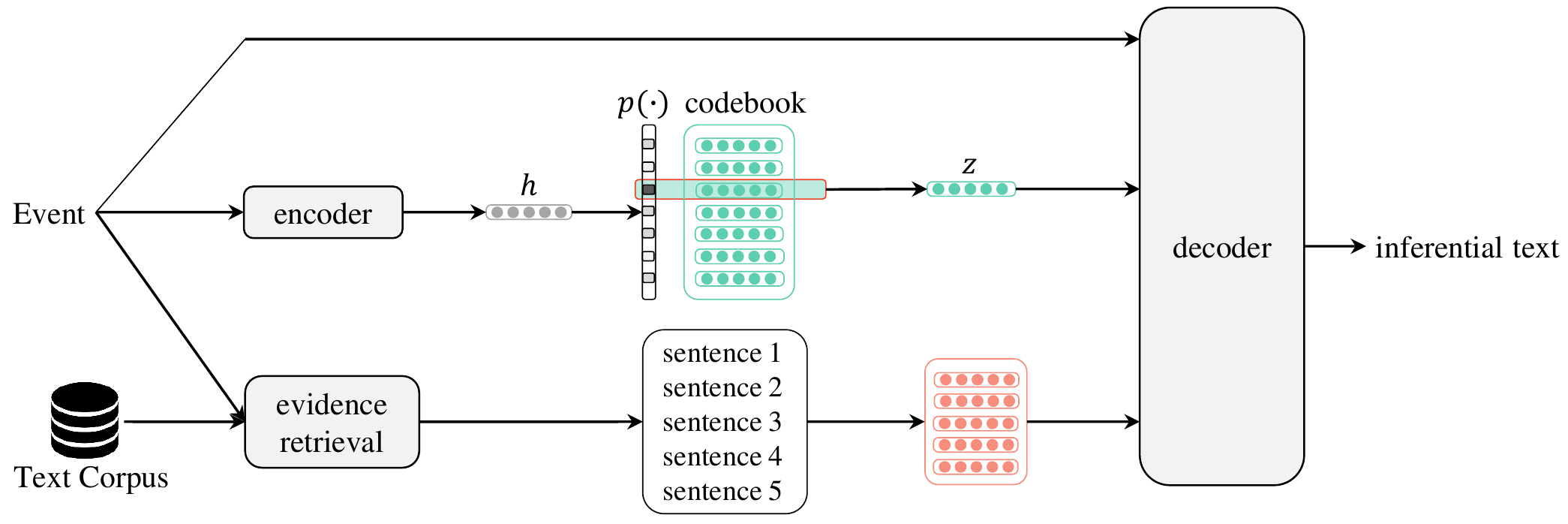}
	\caption{The model architecture of our approach. }
	\label{fig:vq-vae}
\end{figure*}

Figure \ref{fig:vq-vae} illustrates the model architecture of our approach. The model is based on encoder-decoder framework equipped with Vector Quantised-Variational Autoencoder (VQ-VAE) \cite{van2017neural}, where the VQ-VAE is learned to model the latent semantic distribution within inferences given an event. Latent variables $z$ from the VQ-VAE will be used to calculate the \mbox{relevant} of retrieved evidence in the semantic space to guide the generation.

%Thus,  
%The representation captures the latent semantic within targets given a input, 
%which will be used to automatically select relevant evidence in a semantic space to guide the generation. 

Compared with continuous VAEs, VQ-VAE does not suffer from ``posterior collapse'' issues that latent variables are often ignored with a powerful decoder \cite{van2017neural}.
VQ-VAE mainly consists of three parts: a codebook for modeling the latent semantic distribution within inferences over discrete latent variables, a recognition network for modeling a posterior distribution $q_\phi(z|x,y)$, and a prior network for inferring a prior distribution $p_\theta(z|x)$.

\paragraph{Codebook}
A codebook aims to model the \mbox{latent} semantic discrete distribution within inferences, which is composed of $k$ discrete latent variables (i.e. $k$-way  categorical). We define the codebook as an embedding table $T\in{R^{k\times d}}$, where $d$ is the dimension of latent variables. The semantic latent variable $z$ is indexed from the posterior distribution $q_\phi(z|x,y)$ in the training phase and the prior distribution $p_\theta(z|x)$ in the inference phase over the codebook, respectively. 

\paragraph{Posterior Distribution} We follow \citet{van2017neural} to model a discrete posterior distribution $q_\phi(z|x,y)$ over the codebook.
First, we use Transformer \cite{vaswani2017attention} with two layers as our encoder, where the input sequence is the concatenation of an event $x$ and its inference $y$. In order to obtain the representation of an example $(x,y)$, we add a special token in the last of the input sequence and take the hidden state $h_{(x,y)}$ of the special token as the representation of the example. 
The posterior categorical probability distribution $q_\phi(z|x,y)$ is defined as one-hot as follows.
\begin{equation}\label{p_phi(z_k|x,y)}
q_\phi(z_k|x,y)=\left\{
\begin{aligned}
& 1\quad\!\!\! if\ k=\mathop{\arg\min}_{j}||h_{(x,y)}-z_j||_2 \\
& 0\quad\!\!\! otherwise \\
\end{aligned}
\right.
\end{equation}
As we can see, the hidden state $h_{(x,y)}$ of the example is mapped onto the nearest element $z'$ of the codebook under the posterior distribution $q_\phi(z|x,y)$.
\begin{equation}\label{z'}
z'=z_k\quad where\  k=\mathop{\arg\min}_{j}||h_{(x,y)}-z_j||_2 
\end{equation}

\paragraph{Prior Distribution}
In the inference phase, only the event $x$ is given, which requires a prior distribution estimator to infer the prior distribution $p_\theta(z|x)$. Since the prior distribution is crucial for the inference phase, we use a powerful pre-trained language model such as RoBERTa \cite{liu2019roberta} to encode the event into a hidden state $h$. 
Since the prior distribution is categorical, we then use a $k$-way classifier following a softmax function to infer the prior distribution, where $W_k\in{R^{d\times k}}$ is the model parameters.
\begin{equation}
p_\theta(z|x)=softmax(hW_k)
\end{equation}

The training detail of the VQ-VAE will be introduced in the Section \ref{training}.

\subsection{Evidence Retrieval}
In this section, we describe how to retrieve event-related evidence as background knowledge. Given an event, we expect that retrieved evidence can contain the event and provide its context as a clue to guide the generation. 

To retrieve event-related evidence, we use the event as a query to search evidence from a large \mbox{text} corpus. Specifically, we first remove stop words in the given event and then concatenate the \mbox{words} as a query to search evidence from the corpus by \mbox{Elastic} Search engine\footnote{\url{https://www.elastic.co/}}. The engine ranks the matching scores between the query and all sentences using BM25 and select top $K$ sentences as evidence $C=\{c_1,c_2,...,c_K\}$.
To provide detailed context about the event, we build our corpus upon BooksCorpus \cite{zhu2015aligning} that consists of 11,038 story books,
since stories usually give a detailed account of an event such as causes and effects of the event.   

\subsection{Evidence-Aware Decoder}
In this section, we propose an evidence-aware decoder, which consists of two components, evidence selection and a generator, respectively. Evidence selection aims to calculate a context distribution $p_s(c|z)$ given a latent variable $z$ to model the relevance of retrieved evidence, while the generator $p_m(y|x,c)$ takes an event $x$ and evidence $c$ as the input to generate the inferential text $y$.

%In this part, we present the way to use a semantic latent variable from VQ-VAE to find relevant evidence. 

%We first remove stopwords in the given event and then concatenate the words as a query to search candidate evidence from a large-scale text corpus by Elastic Search engine\footnote{\url{https://www.elastic.co/}}. The engine ranks the matching scores between the query and all paragraphs using BM25 and select top $K$ paragraphs as evidence $C=\{c_1,c_2,...,c_K\}$. 

\subsubsection{Evidence Selection}
The relevance of retrieved evidence is different depending on the semantics of inference, which requires a context distribution to model the relevance. For examples, given an event \textit{``PersonX reads PersonY's diary''} and its inference ``\textit{PersonX feels guilty}'', the relevance of the evidence ``\textit{PersonX stole PersonY's diary}'' should be higher than that of the evidence   ``\textit{PersonY invites PersonX to read his diary}''.
However, inferences are unseen in the inference phase, thus we cannot use inferences to model the context distribution.
Instead, we \mbox{utilize} semantic latent variables from the VQ-VAE that models the latent semantic distribution of inferences given an event to calculate the relevance of retrieved evidence.

Evidence selection aims to calculate a context distribution $p_s(c|z)$ over retrieved evidence given a semantic latent variable $z$ to model the relevance of retrieved evidence. 
Considering that term-based retrieval (i.e. BM25) may fail to retrieve relevant evidences and all retrieved evidence cannot support the generation, we add an empty evidence $c_\phi$ into the set $C$ of retrieved evidence as the placeholder. We first use Transformer with two layers to encode retrieved evidence into context vectors $H_C=\{h_{c_1},h_{c_2},..,h_{c_K},h_{c_\phi}\}$ in the semantic space.  Then, the context distribution $p_s(c|z)$ over retrieved evidence given the semantic latent variable $z$ is calculated as one-hot as follows.
\begin{equation}\label{p(c_k|z)}
p_s(c_k|z)=\left\{
\begin{aligned}
& 1\quad\!\!\! if\ k=\mathop{\arg\min}_{j}||h_{c_j}-z||_2 \\
& 0\quad\!\!\! otherwise \\
\end{aligned}
\right.
\end{equation}
As we can see, the latent variable $z$ is mapped onto the nearest element $c_{z}$ of the retrieved evidence under the context distribution $p_s(c|z)$.
\begin{equation}\label{c_{z}}
c_{z}=c_k\quad where\  k=\mathop{\arg\min}_{j}||h_{c_j}-z||_2
\end{equation}

Another ``soft'' distribution such as using an attention mechanism to calculate the relevance of retrieved evidence can also model the context distribution, but we choose the one-hot distribution as our context distribution since it maps the latent variable $z$ onto the nearest element of the retrieved evidence, the property of which can help effectively learn the model (described in the Section \ref{training}).

\subsubsection{Generator}
Recently, Transformer-based \cite{vaswani2017attention} language models like GPT-2 \cite{radford2019language} have achieved strong performance in text generation, which is pre-trained from a large-scale text corpus and then fine-tuned on downstream tasks. In this work, we use the GPT-2 $p_m(y|x,c)$ as the backbone of our generator and further take retrieved evidence into account. 

A general approach to utilize evidence to guide the generation is to calculate the context vector $h_c=\sum_{i=1}^{K+1}p_s(c_i|z)h_{c_i}$ as the input of GPT-2 according to the relevance $p_s(c|z)$ of retrieved evidence.
However, this approach changes the architecture of GPT-2, invalidating the original weights of pre-trained GPT-2.
Instead, we sample an evidence $c$ from the context distribution $p_s(c|z)$ and then concatenate the event and the selected evidence as the input. 

To make the paper self-contained, we briefly describe the GPT-2, which takes an evidence and an event as the input and generates the inference $y=\{y_1,y_2,..,y_n\}$. 
This model applies N transformer layers over the input tokens to produce an output distribution over target tokens:
\begin{equation}
\begin{split}
&h^0=[c;x;y_{<t}]W_e+W_p \\
&h^l=transformer_{l-1}(h^{l-1})\\
&p(y_t)=softmax(h^{N-1}_{last}W^T_e) \\
\end{split}
\end{equation}
where $W_e$ is the token embedding matrix, $W_p$ is the position embedding matrix, and $h^{N-1}_{last}$ is the hidden state of the last token on the top layer. Each transformer layer $transformer_{l-1}$ contains an architecturally identical transformer block that applies a masked multi-headed self-attention operation followed by a feed forward layer over the input $h^{l-1}$ in the $l$-th layer.  
\begin{equation}\label{equa:transformer-block}
\begin{split}
&\hat{g}^l=MultiAttn(h^{l-1}) \\
&{g}^l=LN(\hat{g}^l+h^{l-1})\\
&\hat{h}^l=FFN({g}^l)\\
&h^l=LN(\hat{h}^l+{g}^l)\\
\end{split}
\end{equation}
where $MultiAttn$ is a masked multi-headed self-attention mechanism, which is similar to \citet{vaswani2017attention}, $FFN$ is a two layers feed forward network, and $LN$ represents a layer normalization operation \cite{ba2016layer}.

\subsection{Training}
\label{training}
Our entire approach corresponds to the following generative process. Given an event $x$, we first sample a latent variable $z$ from the VQ-VAE $p_\theta(z|x)$. We then select relevant evidence $c$ according to the semantics of the latent variable from the context distribution $p_s(c|z)$. Finally, the generator $p_m(y|x,c)$ takes the event $x$ and the selected evidence $c$ as the input and generate the inference $y$. Therefore, the probability distribution $p(y|x)$ over inferences $y$ given the event $x$ is formulated as follow. 
\begin{equation}
p(y|x)=\sum_{z\in T}\sum_{c\in C}p_m(y|x,c)p_s(c|z)p_\theta(z|x)
\end{equation}

A straightforward method for learning our model might be maximizing the marginal likelihood by joint learning, but it is computationally intractable. 
Instead, we first learn the VQ-VAE with the prior distribution $p_\theta(z|x)$ in isolation, which can enable the codebook to capture the latent semantics within inferences. Then, we train the evidence-aware decoder under the posterior distribution $q_\phi(z|x,y)$.

\paragraph{Training VQ-VAE}
To enable the codebook to capture the latent semantics within inferences, we train the VQ-VAE by reconstructing the inferential text $y$ using the latent variable $z$.  We use the pre-trained language model GPT-2 \cite{radford2019language} as our decoder to generate the inference $p(y|x,z)$, where the input is the sum of token embedding, position embedding and the latent variable $z$. To make reconstruction better conditioned on the latent variable, we replace each query in the multi-head self-attention mechanism with the sum of the latent variable and the query, as well for keys, values and hidden states on the top layer. We follow \citet{van2017neural} to learn the VQ-VAE by minimizing the loss function. 
\begin{equation}
\begin{aligned}
loss_{rec}&=-logp(y|x,h_{(x,y)}+sg[z-h_{(x,y)}])+\\
&\quad  ||sg[h_{(x,y)}]-z||^2_2+\beta||h_{(x,y)}-sg[z]||^2_2
\end{aligned}
\end{equation}
where $sg$ stands for the stop gradient operator that has zero partial derivatives during differentiation, and $\beta$ is a hyperparameter which controls the speed to change the latent variable. We set the $\beta$ as 0.25 in all experiments. The decoder optimizes the first loss term (reconstruction) only, the encoder optimizes the first and the last loss terms, and the codebook are updated by the middle loss term.

We obtain the posterior distribution $q_\phi(z|x,y)$ after optimizing the encoder and the codebook. Afterward, we learn the prior distribution estimator to infer the prior distribution $p_\theta(z|x)$. 
Since the posterior distribution is categorical, we can calculate approximate prior distributions as follow in the training dataset $D$, where $N_{(x)}$ is the number of examples that includes the event $x$. 
\begin{equation}
p(z|x)=\sum_{(x,y_i)\in D}\frac{q_\phi(z|x,y_i)}{N_{(x)}}
\end{equation}

Therefore, we can fit the prior distributions by minimizing the KL divergence.
\begin{equation}
loss_{prior}=KL(p(z|x)||p_\theta(z|x))
\end{equation}

\paragraph{Training Evidence-Aware Decoder}
After training VQ-VAE, we jointly learn the context distribution $p_s(c|z)$ and the generator $p_m(y|x,c)$ by maximizing the following marginal likelihood under the posterior distribution $q_\phi(z|x,y)$.
\begin{equation}\label{p_y_1}
logp(y|x)=E_{z\sim q_\phi}[\sum_{c\in C}logp_m(y|x,c)p_s(c|z)]
\end{equation}

According to the Equation \ref{z'}, the example $(x,y)$ is mapped onto the nearest element $z'$ of the codebook under the posterior distribution $q_\phi(z|x,y)$. Meanwhile, according to the Equation \ref{c_{z}}, the latent variable $z'$ is mapped onto the nearest element $c_{z'}$ of retrieved evidence. 
Therefore, the objective in Equation \ref{p_y_1} can be simplified as follow.

\begin{equation}
logp(y|x)=logp_m(y|x,c_{z'})+logp_s(c_{z'}|z')
\end{equation}

Since the ground truth evidence for the example is unobserved, we cannot directly train the model by maximizing the marginal likelihood. To remedy this problem, we use reinforcement learning algorithm to optimize the objective.
\begin{equation}
\begin{aligned}
R&=\delta (p_m(y|x,c_{z'})-p_m(y|x,c_{r})) \\
logp(y|x)&=logp_m(y|x,c_{z'})+ Rlogp_s(c_{z'}|z') \\
\end{aligned}
\end{equation}
where $R$ is the reward designed to guide the model training, $\delta (x)$ is 1 if $x$ is larger than 0 otherwise $-1$, and $c_{r}$ is a randomly selected evidence where $c_{r}\neq c_{z'}$. 
The idea of designing the reward is that correct evidence should increase the probability of the gold inference compared with other evidence.
Note that there is no real gradient defined for $p_s(c|z)$, instead, we approximate the gradient similar to the straight-through estimator \cite{bengio2013estimating}.
\begin{equation}\label{logp(y|x)}
logp(y|x)=logp_m(y|x,c_{z'})- R||h_{c_{z'}}-z'||^2_2
\end{equation}

%assuming that semantic parser provides the true conditional distribution over the target $y$ given context $c$ and retrieved examples $S$ under the joint distribution $p_{ret}(S|x,c)p_{data}(x,c,y)$. 
%Then, we optimize a lower bound for the marginal likelihood under this semantic parser \cite{hashimoto2018retrieve}, which decomposes the reconstruction term and the KL divergence as follows. 
%\begin{align}\label{equa:KL}
%\notag
%logp(y|x,c)\geq E_{z\sim p(z|x,c)}logp(y|z)  \\
%-E_{\{({x',c')\}} \sim p_{ret}}KL(p(z|x,c)||p(z|x',c'))
%\end{align}  

\begin{table*}[t!]
	\begin{center}
		{\small
			\begin{tabular}{l|cccccccccc} \hline 
				%         		&\multicolumn{9}{c}{Dev} \\
				%         		\hline
				Methods& xIntent &  xNeed & xAttr  & xEffect & xReact & xWant & oEffect& oReact& oWant&Overall\\
				\hline
				\hline
				\multicolumn{11}{c}{Single Task} \\
				\hline
				\hline
				
				S2S & 8.17 &  12.35  & 2.96 & 5.26 &3.43&13.44 &6.42 &4.09 &7.08&7.02\\
				
				VRNMT & 9.52 &  13.35  & 4.87 & 4.42 &7.64 &9.80 &13.71 &5.28&10.79&8.82\\
				
				CWVAE & 12.12 &  15.67  & 5.63 & 14.64 &8.13 &15.01 &11.63 &8.58&13.83&11.69\\
				\hline
				\hline
				\multicolumn{11}{c}{Multi Task} \\
				\hline
				\hline
				S2S* & 24.53 &  23.85  & 5.06 & 9.44 &5.38&24.68 &7.93 &5.60 &21.30&14.20\\
				COMET* & 25.82 &  25.54  & 5.39 & 10.39 &5.36 &\bf{26.41} &8.43 &5.65&21.96&15.00\\	
				COMET & - &  -  & - & - &- &- &- &-&-&15.10\\	
				EA-VQ-VAE & \bf{26.89} &  \bf{25.95}  & \bf{5.72} & \bf{10.96} &\bf{5.68} &25.94 &\bf{8.78} &\bf{6.10}&\bf{22.48}&\bf{15.40}\\	
				\hline
			\end{tabular}
		}
	\end{center}
	\caption{\label{table:ATOMIC_bleu} BLEU score on nine inference dimensions of the ATOMIC test dataset with different approaches. For inference dimensions, ``x'' and ``o'' refers to PersonX and others, respectively (e.g. ``xAttr'': attribute of PersonX, ``oEffect'': effect on others). The tag (*) means re-implementation.}
	\vspace{-0.4cm}
\end{table*}

Thus, we can optimize the evidence-aware decoder by maximizing the marginal likelihood in the Equation \ref{logp(y|x)}. Please see more details about the model hyperparameters in Appendix B.

\section{Experiment}
%For GPT-2, we use the pre-trained GPT-2 \cite{radford2019language} with 12 layers, 768 dimensional hidden states and 12 attention heads.
\subsection{Model Comparisons}
Following \citet{sap2019atomic}, we first use the average BLEU-2 score between each sequence in the top 10 predictions and the gold generations to evaluate the accuracy of generations. We report the result of existing methods on ATOMIC and Event2Mind datasets in the Table \ref{table:ATOMIC_bleu} and Table \ref{table:Event2Mind_bleu}, respectively.

\begin{table}[h]
	\begin{center}
		{\small
			\begin{tabular}{l|cccc} \hline 
%         		&\multicolumn{9}{c}{Dev} \\
%         		\hline
				Methods& xIntent & xReact & oReact &Overall\\
         		\hline
         		\hline
         		\multicolumn{5}{c}{Single Task} \\
         		\hline
         		\hline
				S2S & 2.75 &  2.11  & 5.18&3.35\\
				VRNMT& 4.81 &3.94&6.61&4.03 \\
				CWVAE & 12.98 &  5.65  & 6.97 & 8.53\\
         		\hline
         		\hline
         		\multicolumn{5}{c}{Multi Task} \\
         		\hline
         		\hline
				S2S* & 19.18 &  4.81  & 4.29&9.43\\
				COMET* & 21.64 &  5.10  & 4.36 & 10.37\\	
				EA-VQ-VAE& \bf{23.39} &  \bf{5.74}  & \bf{4.81}&\bf{11.31}\\	
				\hline
			\end{tabular}
		}
	\end{center}
	\caption{\label{table:Event2Mind_bleu} BLEU score on three inference dimensions of the Event2Mind test dataset with different approaches. For inference dimensions, ``x'' and ``o'' refers to PersonX and others, respectively. The tag (*) means re-implementation. }
	\vspace{-0.2cm}
\end{table}

These approaches are divided into two \mbox{groups}. The first group trains distinct models for each inference dimension separately, while the second group trains a model in a multi-task learning way for all inference dimensions. 
{\bf S2S} is a RNN-based  sequence-to-sequence model \cite{sutskever2014sequence}. 
{\bf VRNMT} \cite{su2018variational} introduces a sequence of recurrent latent variables to model the semantic distribution of inferences.
{\bf CWVAE} propose a context-aware variational autoencoder \cite{du2019modeling} to acquire context information, which is first pre-trained on the auxiliary dataset and then fine-tuned for each inference dimension. 
{\bf COMET} \cite{Bosselut2019COMETCT} concatenate the event with an inference dimension as the input and fine-tune the pre-trained GPT-2. Since COMET does not report the performance for each inference dimension, we re-implement the model for better comparison. Our approach is abbreviated as {\bf EA-VQ-VAE}, short for Evidence-Aware Vector Quantised Variational AutoEncoder.

As we can see in the Table \ref{table:ATOMIC_bleu} and Table \ref{table:Event2Mind_bleu}, the multi-task learning performs better than single-task learning overall. Therefore, we train our model in a multi-task way and compare our approach with multi-task learning based methods. 
From the Table \ref{table:ATOMIC_bleu}, we can see that our approach \mbox{performs} better on the majority of inference dimensions, achieving the state-of-the-art result on ATOMIC dataset. For the Event2Mind dataset, results in the Table \ref{table:Event2Mind_bleu} show that our approach brings a gain of 1\% BLEU score overall compared with the state-of-the-art method.
%From the Table \ref{table:ATOMIC_bleu}, we can see that our model exceeds the performance of other methods, achieving a gain of 1.1\% BLEU score overall compared with the state-of-the-art method on the Event2Mind dataset. For the ATOMIC dataset with broader inferential dimensions, results in the Table \ref{table:ATOMIC_bleu} also show that our approach performs better on the majority of inference dimensions.

 \begin{table}[h]\small
	\centering
	\begin{tabular}{l|c|c|c|c}
		\hline
		\multirow{2}{*}{Methods}& \multicolumn{2}{c|}{Event2Mind} & \multicolumn{2}{c}{ATOMIC} \\
		\cline{2-5}
		& dist-1 & dist-2 & dist-1 & dist-2 \\
		\hline
		%		\hline
		S2S* & 638 &  1,103  & 2,193&5,761 \\
		COMET* & 1,794 &  4,461  & 3,629 & 12,826 \\
		EA-VQ-VAE & \bf{1,942} &  \bf{4,679}  & \bf{3,918} & \bf{14,278} \\ 
		\hline
	\end{tabular}
	\caption{The number of distinct n-gram (dist-1 and dist-2) overall on Event2Mind and ATOMIC test dataset with different multi-task learning based methods. The tag (*) means re-implementation.}
	\label{table:diversity} 
	\vspace{-0.2cm}
\end{table}

Besides, in order to evaluate the diversity of generations, we use the number of distinct unigrams (dist-1) and bigrams (dist-2) as evaluation \mbox{metrics} \cite{li2015diversity}. Since we train our model in a multi-task way, we compare our approach with multi-task learning based methods for fair comparison. Results in the Table \ref{table:diversity} show that our approach could increase the diversity of generations overall on both datasets.

Since automatic evaluation of generated language is limited \cite{liu-etal-2016-evaluate}, we also perform a human evaluation on model performance. Following the setup of \cite{sap2019atomic}, we evaluate 100 randomly selected examples from the test set and use beam search to generate 10 candidates from different models. Five human experts are asked to identify whether a model generation is correct given an event with an inference dimension. Table \ref{human} shows the result of the human evaluation on both datasets, where our approach achieves a gain of 1.5\%$\sim$2\% accuracy compared with {\bf COMET}.
%Overall, we obtain 5000 rating for each model (100 examples $\times$ 10 candidates $\times$ 5 experts). 

 \begin{table}[h]\small
	\centering
	\begin{tabular}{l|c|c}
		\hline
		Methods & Event2Mind &  ATOMIC \\
		\hline
		%		\hline
		S2S* & 0.3901 &  0.5174  \\
		COMET* & 0.4874 &  0.6379  \\
		EA-VQ-VAE & \bf{0.5072} &  \bf{0.6528}   \\ 
		\hline
	\end{tabular}
	\caption{Human score (accuracy) of generations on Event2Mind and ATOMIC test dataset. The tag (*) means re-implementation.}
	\label{human} 
	\vspace{-0.2cm}
\end{table}

\subsection{Model Analysis}
We conduct ablation analysis to better understand how various components in our approach impact overall performance. We remove evidence and VQ-VAE, respectively, to analyze their contribution. 
\begin{table}[h]
	\begin{center}
		{\small
			\begin{tabular}{l|cccc} \hline 
				%         		&\multicolumn{9}{c}{Dev} \\
				%         		\hline
				Methods& xIntent & xReact & oReact &Overall\\
				\hline
				EA-VQ-VAE & 23.37 &  5.83  & 4.87&11.32\\
				\ - w/o evidence & 21.69 &  5.36  & 4.48&10.51 \\
				\ - w/o VQ-VAE & 21.87 &  5.41  & 4.60&10.63 \\
				\ - w/o SL & 21.95 & 5.54  & 4.57&10.69 \\
				\hline
			\end{tabular}
		}
	\end{center}
	\caption{\label{table:analysis} BLEU score on the Event2Mind dev dataset with different approaches. SL is short for separately learning.}
\end{table}

Table \ref{table:analysis} shows that the overall performance drops from 11.3\% to 10.5\% on Event2Mind dev dataset when removing the evidence totally (w/o evidence), which reveals the importance of evidence for inferential texts generation. 
After ablating the VQ-VAE and selecting top-1 evidence as background (w/o VQ-VAE), we can see that the performance \mbox{drops} from 11.3\% to 10.6\%, which means VQ-VAE can automatically select relevant and useful evidence. 
In order to demonstrate the effectiveness of our learning method, we also train our model by joint learning (w/o SL). The overall BLEU score drops from 11.3\% to 10.7\%, which shows that our learning method can effectively train our model. 
\begin{figure}[h]
	\centering
	\includegraphics[width=.47\textwidth]{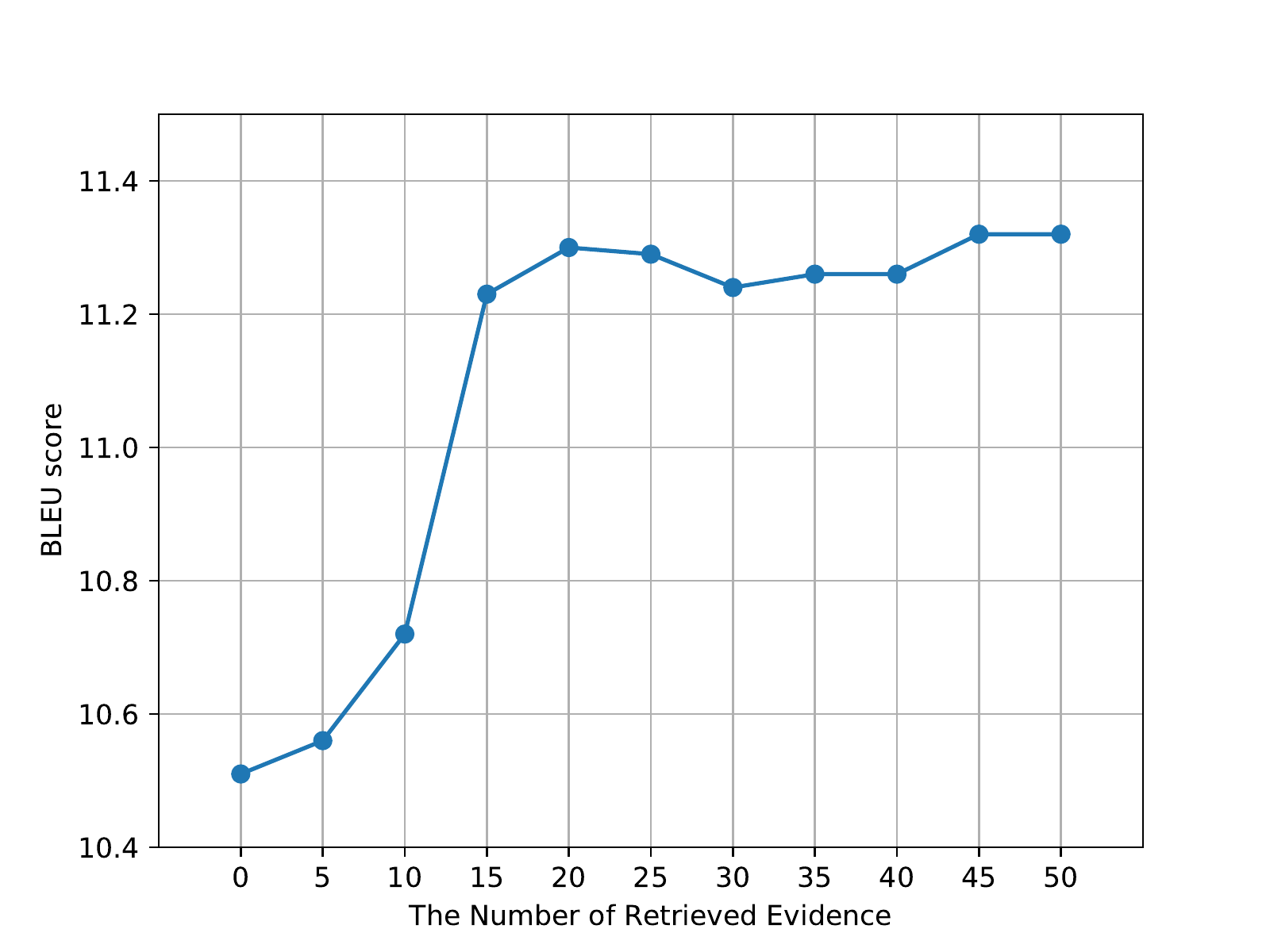}
	\caption{Overall performance with different number of retrieved evidence on Event2Mind dev dataset.}
	\label{fig:k}
\end{figure}

We also study how the amount of evidence retrieved from the corpus impacts the performance. From Figure \ref{fig:k}, we can see that overall BLEU \mbox{score} increases as the number of retrieved evidence expands.
This is consistent with our intuition that the performance of our approach is improved by expanding retrieved examples, since our approach can select relevant and useful evidence from more retrieved evidence. When the number of retrieved evidence is larger than 20, the overall performance does not improve. The main reason is that the quality and relevance of retrieved evidence decreases as the number of retrieved evidence expands.

%In order to demonstrate the importance of automatically selecting relevant evidence by the VQ-VAE, we remove the VQ-VAE and select an evidence with top-1 matching scores using BM25 to guide the generation (w/o VQ-VAE). We can see that 

\begin{figure*}[t!]
	\centering
	\includegraphics[width=0.96\textwidth]{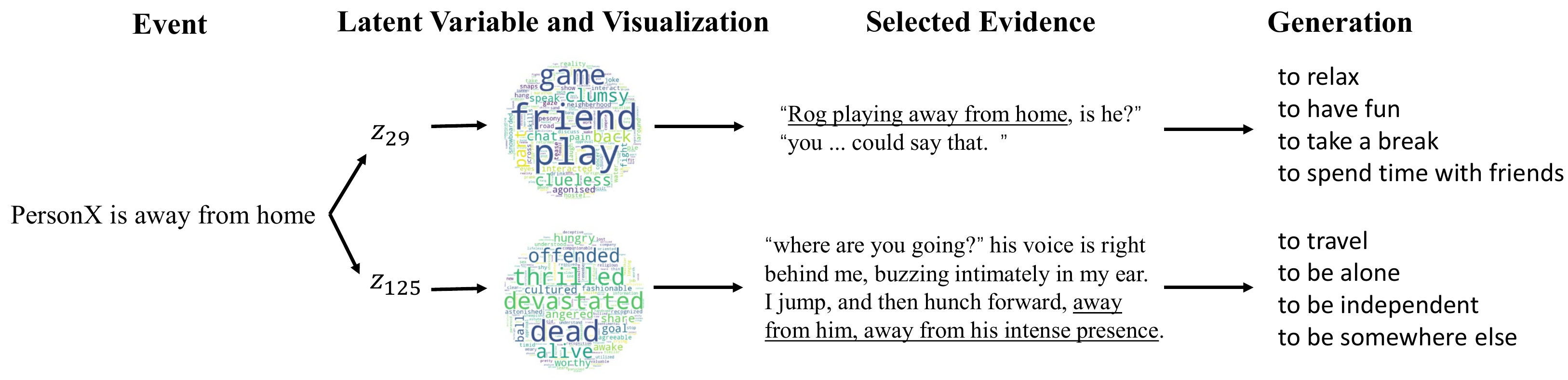}
	\caption{An examples of Event2Mind dataset on the xIntent dimension (i.e. \textit{``PersonX wants''}).}
	\label{fig:case_study}
\end{figure*}

\subsection{Case Study}
We give a case study to illustrate the entire procedure of our approach. Figure \ref{fig:case_study} provides an example of the generations given an event \textit{``PresonX is away from home''} on the ``xIntent'' dimension (i.e. \textit{``PersonX wants''}). We first sample two latent variables from the codebook (i.e. $z_{29}$ and $z_{125}$) according to the prior distribution of VQ-VAE. We visualize the semantics of latent variables by displaying word cloud of examples that are under the same latent assignment. As we can see, $z_{29}$ captures the positive semantics like \textit{``play''} and \textit{``friend''}, while $z_{125}$ captures the negative semantics like \textit{``devastated''} and \textit{``offended''}. Then, two latent variables are respectively used to select relevant evidence as background knowledge. As we can see, the first latent variable selects an evidence about \textit{``playing''}, which provides a clue for the model to generate texts such as \textit{``to have fun''} and \textit{``to spend time with friends''}. Another latent variable selects another evidence in a quarrel scene, which can help the model reason about \textit{``PersonX wants to be alone''}. The case study shows that our approach not only equips the generation with an explicit control over the semantics of evidence but select relevant evidence to guide the generation. Please find another case on other inference dimension on Appendix C.

\subsection{Error Analysis}
We analyze 100 incorrectly predicted instances randomly selected from the ATOMIC dataset, and summary two main classes of errors.
The first problem is that some examples cannot retrieve relevant evidence since the scale of text corpus is limited. We can leverage more sources like Wikipedia to retrieve evidence. 
Another cause of this problem is that term-based retrieval (e.g. BM25) calculates the matching score using words overlap and cannot capture semantics of sentences. For examples, the evidence\textit{``the lights began to shift away from the fire, like a line of fireflies''} will be retrieved for the event  \textit{``PersonX lights a fire''} since of the high overlap, but the event does not occur in the evidence.
This problem might be mitigated by using better semantic-based retrieval model.
%Therefore, we can improve performance of the retrieval by better semantic-based retrieval model.
The second problem is that the model cannot effectively leverage selected evidence. Although the selected evidence is closely related to the event and the inference can be obtained from the evidence, the model still generate incorrect texts since lacking of supervised information.  
A potential direction to mitigate the problem is to annotate background knowledge of events in the training dataset. 
%This problem might be mitigated by incorporating a dedicated module such as the attention mechanism that takes selected evidence into account in the decoder.

%Lastly, some examples fails to generate correct texts since ignoring retrieved evidence.

\section{Related Work}

\subsection{Event-Related Text Understanding}
Recently, event-related text understanding has attracted much attention \cite{chambers2008unsupervised,segers2016event,wang2017integrating,li2018constructing,rashkin2018event2mind,sap2019atomic, guo2020inferential}, which is crucial to artificial intelligence systems for automated commonsense reasoning.
There are a variety of tasks that focus on event-related text understanding in different forms.
Script \cite{schank1977scripts} uses a line to represent temporal and causal relations between events, and 
the task of script event prediction \cite{chambers2008unsupervised} requires models to predict the subsequent event given an event context. Previous works on the task are mainly based on event pairs \cite{chambers2008unsupervised,granroth2016happens}, event chains \cite{wang2017integrating}, and event evolutionary graph \cite{li2018constructing} to predict script event. 
In addition, our task relates to story ending prediction \cite{sharma2018tackling,mostafazadeh2016corpus,zellers2018swag}.
\citet{mostafazadeh2016corpus} introduce a dataset for story ending prediction, which requires models to choose the most sensible ending given a paragraph as context.
%\citet{zellers2018swag} constructs SWAG and provides a task like natural language inference, with the requirement for commonsense reasoning.
 %which requires models to choose the most sensible ending given a paragraph as context.
%There are multiple question answering datasets, in which answering questions requires commonsense knowledge involved in the text or extracted from external resources.
%Notably, \newcite{mostafazadeh2016corpus} introduce a dataset for story ending prediction given a paragraph as context.
%\newcite{zellers2018swag} construct SWAG, and provide a task like natural language inference, with the requirement for commonsense reasoning.
In this work, we study inferential text generation proposed by \citet{rashkin2018event2mind} and \citet{sap2019atomic}, both of which focus on generating texts about causes and effects of events and mental \mbox{states} of event participants. 

\subsection{Variational Autoencoder Based Text Generation}
Natural Language Generation, also known as text
generation \cite{mckeown1992text,sutskever2011generating}, has recently become popular in NLP community \cite{feng2018topic,duan2020pretrain}. Recently, Variational Autoencoder (VAE) \cite{kingma2013auto} has achieved promising performance on various text generation tasks, including machine translation \cite{zhang2016variational,su2018variational}, text summarization \cite{miao2016language,li2017deep}, and dialogue generation \cite{serban2017hierarchical,zhao2017learning}. 
%For text generation, \citet{wang2019topic} utilizes the latent variable to model topic distributions in text generation.
For machine translation, \citet{zhang2016variational} and \citet{su2018variational} introduce a continuous latent variable to explicitly model the semantics of a source sentence, which is used to guide the translation. In dialogue genration, \citet{serban2017hierarchical} apply a latent variable hierarchical encoder-decoder model to facilitate longer response, while \citet{zhao2017learning} uses latent variables to capture potential conversational intents and generates diverse responses.  
A recent work CWVAE \cite{du2019modeling} on event-centered If-Then reasoning is the most related to our work, which introduces an additional context-aware \mbox{latent} variable to implicitly guide the generation by a two-stage training procedure. 
Different with previous works, we introduce a discrete latent variable to capture underlying semantics within inferences based on VQ-VAE that does not suffer from ``posterior collapse'' issues \cite{van2017neural}. These discrete latent variables are used to selectively leverage evidence as background knowledge to explicitly guide the generation. Besides, our approach provides a way to uncover the rationale of a generation to some extent through tracing back the evidence that supports the generation and the selected discrete latent variable.

%to capture underlying semantic within inferences
\section{Conclusion}
In this paper, we present an evidence-aware generative model based on VQ-VAE, which utilizes discrete semantic latent variables to select evidence as background knowledge to guide the generation. Experimental results show that our approach achieves state-of-the-art performance on Event2Mind and ATOMIC datasets. Further analysis shows that our approach selectively uses evidence to generate different inferential texts from multiple perspectives. 

\section*{Acknowledgments}
Daya Guo and Jian Yin are supported by the National Natural Science Foundation of China (U1711262, U1611264, U1711261, U1811261, U1811264, U1911203), National Key R\&D Program of China (2018YFB1004404), Guangdong Basic and Applied Basic Research Foundation (2019B1515130001), Key R\&D Program of Guangdong Province (2018B010107005). Jian Yin is the corresponding author.

\bibliography{anthology,acl2020}
\bibliographystyle{acl_natbib}

\appendix

\begin{table*}[t]
	\begin{center}
		{\small
			\begin{tabular}{c|c|l|l} 
				\hline 
				Event&Inference dim &\multicolumn{1}{c|}{Description}& \multicolumn{1}{c}{Target} \\
				\hline
				\multirow{9}{*}{PersonX runs away from home}&\multirow{3}{*}{xIntent}&\multirow{3}{*}{because PersonX wanted to}&to leave his home,  \\
				~&~&~&to be independent,  \\
				~&~&~&be away from a parent \\
				\cline{2-4}
				~&\multirow{3}{*}{xReact}&\multirow{3}{*}{as a result, PersonX feels}&lonely,  \\
				~&~&~&nervous,  \\
				~&~&~&regretful \\
				\cline{2-4}
				~&\multirow{3}{*}{oReact}&\multirow{3}{*}{as a result, others feel}&sad,  \\
				~&~&~&angry, \\
				~&~&~&worried  \\
				\hline 
			\end{tabular}
		}
	\end{center}
	\caption{\label{table:Event2Mind} Examples of Event2Mind dataset, including three inference dimensions. For inference dimensions, ``x'' and ``o'' refers to PersonX and others, respectively (e.g. description of ``xIntent'': \textit{Because PersonX wants}). }
\end{table*}

\begin{table*}[t]
	\begin{center}
		{\small
			\begin{tabular}{c|c|l|l} 
				\hline 
				Event&Inference dim &\multicolumn{1}{c|}{Description}& \multicolumn{1}{c}{Target} \\
				\hline
				\multirow{18}{*}{PersonX visits friends}&\multirow{2}{*}{xIntent}&\multirow{2}{*}{because PersonX wanted to}&to enjoy their time,  \\
				~&~&~&to catch up with them  \\
				\cline{2-4}
				~&\multirow{2}{*}{xNeed}&\multirow{2}{*}{before that, PersonX needed to}&to go to their location,  \\
				~&~&~&to call them  \\
				\cline{2-4}
				~&\multirow{2}{*}{xAttr}&\multirow{2}{*}{PersonX is seen as}&friendly,  \\
				~&~&~&sociable  \\
				\cline{2-4}
				~&\multirow{2}{*}{xEffect}&\multirow{2}{*}{has an effect on PersonX}&have a nice party,  \\
				~&~&~&have good dinner  \\
				\cline{2-4}
				~&\multirow{2}{*}{xWant}&\multirow{2}{*}{as a result, PersonX wants}&have fun,  \\
				~&~&~&enjoy and spend time  \\
				\cline{2-4}	
				~&\multirow{2}{*}{xReact}&\multirow{2}{*}{as a result, PersonX feels}&happy,  \\
				~&~&~&comfortable  \\
				\cline{2-4}	
				~&\multirow{2}{*}{oReact}&\multirow{2}{*}{as a result, others feel}&happy,  \\
				~&~&~&pleased  \\
				\cline{2-4}	
				~&\multirow{2}{*}{oWant}&\multirow{2}{*}{as a result, others want}&to wind down,  \\
				~&~&~&to clean their home  \\
				\cline{2-4}	
				~&\multirow{2}{*}{oEffect}&\multirow{2}{*}{has an effect on others}&make the relation stronger,  \\
				~&~&~&bring a guest into their home  \\
				\cline{2-4}				
				\hline 
			\end{tabular}
		}
	\end{center}
	\caption{\label{table:ATOMIC} Examples of ATOMIC dataset, including nine inference dimensions. For inference dimensions, ``x'' and ``o'' refers to PersonX and others, respectively (e.g. description of ``xIntent'': \textit{Because PersonX wants})..}
\end{table*}

\begin{table*}[t]\small
	\begin{center}
		{
			\begin{tabular}{l|c|c|c|c|c|c} \hline 
				%         		&\multicolumn{9}{c}{Dev} \\
				%         		\hline
				Dataset& \makecell[c]{\# inference \\dimension} & \makecell[c]{\# unique \\events} & \makecell[c]{\# average words \\of events} &\makecell[c]{\# inferences \\per example} & \makecell[c]{\# dist-1 \\of inferences}& \makecell[c]{\# dist-2 \\of inferences}\\
				\hline
				Event2Mind & 3 &  24716  &5.1 &2.6&10,929&52,830\\
				\hline
				ATOMIC & 9 &  24313  & 5.2&3.6&27,169&20,5659 \\
				\hline
			\end{tabular}
		}
	\end{center}
	\caption{\label{table:statistic} Statistic of Event2Mind and ATOMIC Dataset.}
\end{table*}

\section{Dataset Details}

We show examples of Event2Mind \cite{rashkin2018event2mind} and ATOMIC \cite{sap2019atomic} dataset in Table \ref{table:Event2Mind} and Table \ref{table:ATOMIC}, respectively. The task \mbox{aims} to generate multiple inferential texts given an event with an inference dimension. 
Table \ref{table:statistic} lists statistics of Event2Mind and ATOMIC dataset. Both datasets contain about 25,000 unique events (\# unique events) extracted multiple data sources, where the events has 5 words on average (\# average words of events). Event2Mind focuses on three inference dimensions shown in Table \ref{table:Event2Mind} and  contains about 2.6 inferences on average, while ATOMIC focuses on nine inference dimensions shown in Table \ref{table:ATOMIC} and contains about 3.6 inferences on average. Beside, we list the number of distinct unigram (\# dist-1 of inferences) and bigram (\# dist-2 of inferences) to evaluate the diversity of inferences.
%Given an event like \textit{``PersonX runs away from home''} and an inference dimension such as \textit{``xIntent''}, the task aims to generate multiple inferential texts such as  \textit{``PersonX''}  wants \textit{``to leave his home''},\textit{``to be independent''} and \textit{``away from a parent''}. 

%Event2Mind focuses on three inference dimensions related to mental states of participants, while ATOMIC has nine inference dimensions including mental states, probable pre- and post conditions of the event, and persona status. 

\section{Model Training}
The text corpus is built upon BooksCorpus \cite{zhu2015aligning}. We extract about 24.2M paragraphs from the corpus, where a paragraph has about 50 words. We retrieve 45 evidence from the corpus for all experiments. We initialize GPT-2 with 12 layers, 768 dimensional hidden states and 12 attention heads using the original pre-trained weights \cite{radford2019language}. For VQ-VAE, the codebook is composed of 400 discrete latent variables and the dimension of latent variable is 768. We set the max length of evidence, events and inferences as 64, 64, and 32, respectively.
Model parameters except GPT-2 are initialized with uniform distribution. We use the Adam optimizer to update model parameters. The learning rate and the batch size is set as 5e-5 and 64, respectively.  In the multi-task learning way, we concatenate events and special tokens of inference dimensions as the input to guide the generation in different dimension. We tune hyperparameters and perform early stopping on the development set.

\section{Additional Case Study}
\begin{figure*}[t]
	\centering
	\includegraphics[width=1\textwidth]{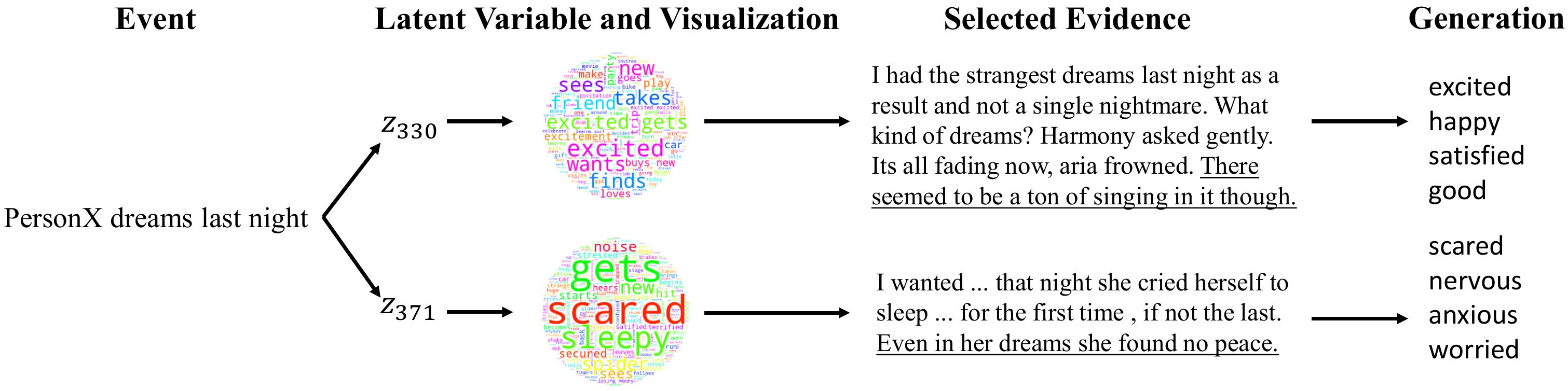}
	\caption{An examples of Event2Mind dataset on the xReact dimension (i.e. \textit{``PersonX feels''}).}
	\label{fig:case_study_1}
\end{figure*}
Figure \ref{fig:case_study_1} provides an example of the generations given an event \textit{``PerxonX dreams last night''} on the ``xReact'' dimension (i.e. \textit{``PersonX feels''}). We first sample two latent variables from the codebook (i.e. $z_{330}$ and $z_{371}$) according to the prior distribution of VQ-VAE \cite{van2017neural}. We visualize the semantics of latent variables by displaying word cloud of examples that are under the same latent assignment. As we can see, $z_{330}$ captures the positive semantics like \textit{``excitied''} and \textit{``friend''}, while $z_{371}$ captures the negative semantics like \textit{``scared''} and \textit{``noise''}. Then, two latent variables are respectively used to select relevant evidence as background knowledge. As we can see, the first latent variable selects an evidence about a sweet dream \textit{``There seems to be a ton of singing in it though''}, which provides a clue for the model to generate positive emotion such as \textit{``excited''} and \textit{``happy''}. Another latent variable select another evidence in a nightmare \textit{``Even in her dreams she found no peace''}, which can help the model reason about the emotion of \textit{``PersonX''} such as \textit{``scared''} and \textit{``nervous''}.

\end{document}